\documentclass{IEEEoj}
\usepackage{cite}
\usepackage{algorithmic}
\usepackage{booktabs} 
\usepackage{tabularx}
\usepackage{algorithm}
\usepackage{multirow} 
\usepackage{pifont}
\usepackage{array} % Ensure this package is included in your document preamble
\usepackage{amsmath,amssymb,amsfonts}
\usepackage{algorithmic}
\usepackage{graphicx,color}
\usepackage{textcomp}
\newcounter{mycounter}
\usepackage{amssymb}
\def\BibTeX{{\rm B\kern-.05em{\sc i\kern-.025em b}\kern-.08em
    T\kern-.1667em\lower.7ex\hbox{E}\kern-.125emX}}
\AtBeginDocument{\definecolor{ojcolor}{cmyk}{0.93,0.59,0.15,0.02}}
\def\OJlogo{\vspace{-14pt}\includegraphics[height=28pt]{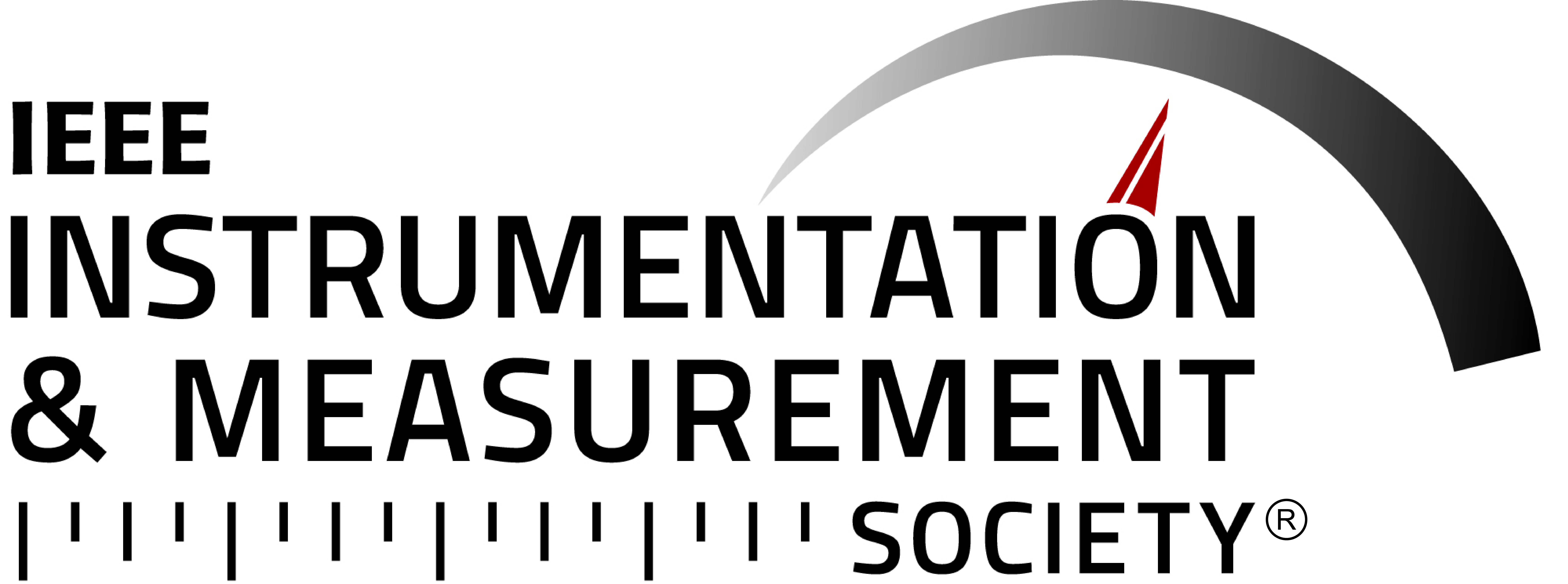}}
\begin{document}
\receiveddate{XX Month, XXXX}
\reviseddate{XX Month, XXXX}
\accepteddate{XX Month, XXXX}
\publisheddate{XX Month, XXXX}
\currentdate{XX Month, XXXX}
\doiinfo{OJIM.2022.1234567}

\newcommand{\cmark}{\includegraphics[height=1.5ex]{checkmark}} % Replace "checkmark" with the actual filename or use a checkmark symbol provided by your LaTeX distribution
\newcommand{\xmark}{\includegraphics[height=1.5ex]{xmark}}       % Replace "xmark" with the actual filename or use a cross symbol provided by your LaTeX distribution

\title{Training and Serving System of Foundation Models: A Comprehensive Survey}

\author{JIAHANG ZHOU\textsuperscript{1}, YANYU CHEN\textsuperscript{1}, ZICONG HONG\textsuperscript{3}, WUHUI CHEN\textsuperscript{2, 4}, YUE YU\textsuperscript{4}, \\ TAO ZHANG\textsuperscript{1}, HUI WANG\textsuperscript{4}, CHUANFU ZHANG\textsuperscript{1}, AND ZIBIN ZHENG\textsuperscript{2}.}

\affil{School of Systems Science and Engineering, Sun Yat-sen University, Guangzhou, 528406, China}
\affil{School of Software Engineering, Sun Yat-sen University, Guangzhou, 510275, China}
\affil{Department of Computing in The Hong Kong Polytechnic University, Hong Kong, 999077, China}
\affil{Peng Cheng Laboratory, Shenzhen, 518000, China}

\corresp{CORRESPONDING AUTHOR: Wuhui Chen (e-mail: chenwuh@mail.sysu.edu.cn).}
% \authornote{This work was supported by the Natural Sciences and Engineering Research Council (NSERC) of Canada.}
\markboth{Preparation of Papers for IEEE OPEN JOURNALS}{Author \textit{et al.}}

\begin{abstract}
Foundation models (e.g., ChatGPT, DALL-E, PengCheng Mind, PanGu-{$\Sigma$}) have demonstrated extraordinary performance in key technological areas, such as natural language processing and visual recognition, and have become the mainstream trend of artificial general intelligence. This has led more and more major technology giants to dedicate significant human and financial resources to actively develop their foundation model systems, which drives continuous growth of these models' parameters. As a result, the training and serving of these models have posed significant challenges, including substantial computing power, memory consumption, bandwidth demands, etc. Therefore, employing efficient training and serving strategies becomes particularly crucial. Many researchers have actively explored and proposed effective methods. So, a comprehensive survey of them is essential for system developers and researchers. This paper extensively explores the methods employed in training and serving foundation models from various perspectives. It provides a detailed categorization of these state-of-the-art methods, including finer aspects such as network, computing, and storage.
Additionally, the paper summarizes the challenges and presents a perspective on the future development direction of foundation model systems. Through comprehensive discussion and analysis, it hopes to provide a solid theoretical basis and practical guidance for future research and applications, promoting continuous innovation and development in foundation model systems.
\end{abstract}



\maketitle

\section{INTRODUCTION}
\IEEEPARstart{T}he combination of deep learning techniques and powerful computational capabilities continuously drives the development of artificial general intelligence, ushering us into the era of foundation models. However, achieving successful applications of foundation models is inseparable from comprehensive support at the system level. A foundation model system is built upon extensive training data, state-of-the-art models, high-performance computing resources, and meticulously optimized training and serving algorithms. The primary purpose of this system is to handle complex tasks with heightened precision, such as GPT3\cite{gpt3}, LLaMA\cite{llama}, PanGu-{$\Sigma$}\cite{pangu}, PengCheng Mind\cite{yn} etc.

\begin{figure}[t]
  \centering
  \includegraphics[width=.5\textwidth]{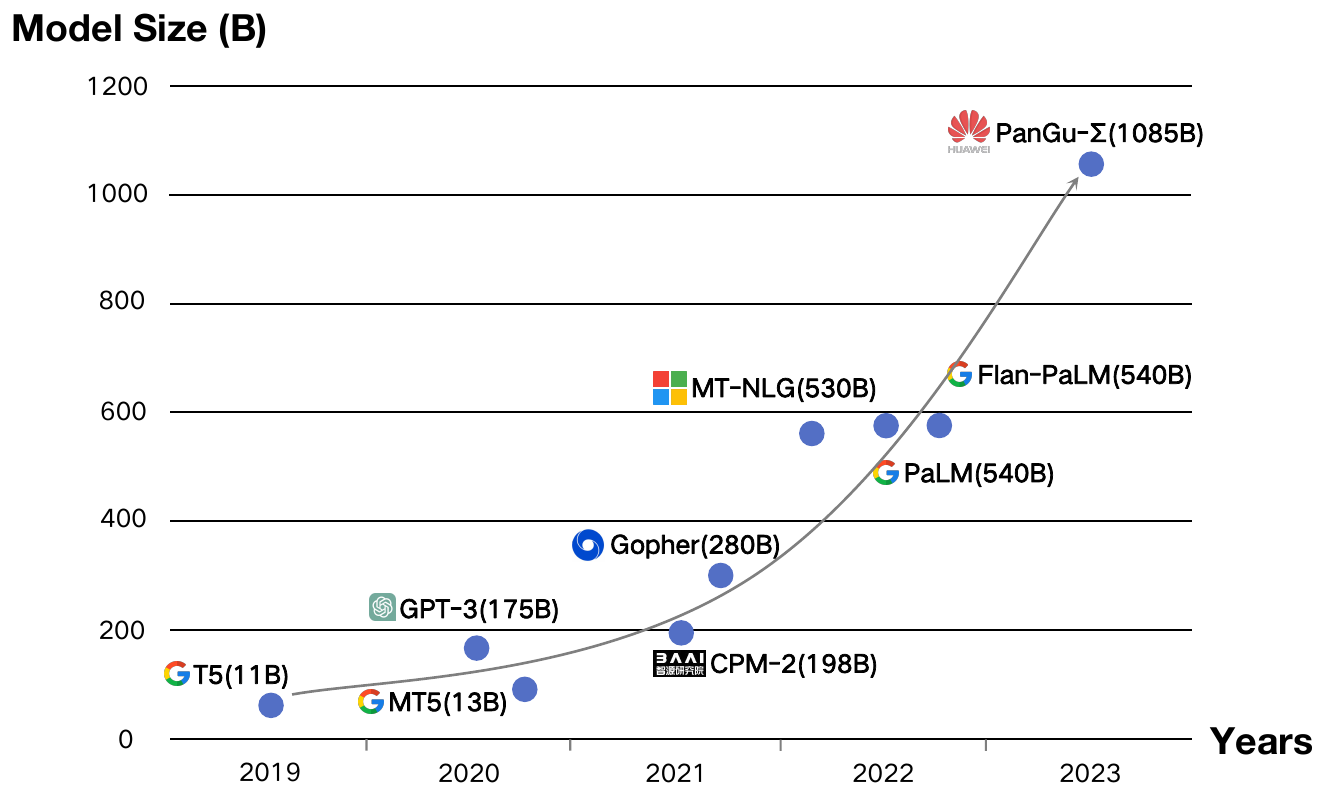}
  \caption{Evolutionary Chart of Model Sizes Over Time.} %caption是图片的标题
  \label{img1}   
\end{figure}

Foundation models have demonstrated extraordinary performance in many tasks. This has led more and more major technology giants to dedicate significant human and financial resources to actively develop their foundation model systems, which increases the parameter size (Figure \ref{img1}).
However, as the parameter size of foundational model systems continues to grow, challenges are posed throughout the lifecycle of foundation models, particularly during the training and serving phases. In the training phase, the substantial parameter size results in significant demands for computation and storage, creating immense pressure on hardware resources and computational efficiency. Consequently, training these models usually takes a long time and requires efficient utilization of computational resources. In the serving phase, with the widespread application of foundation models, the significant increase in workload has become an unavoidable challenge. This heightened demand may lead to issues for serving systems, such as latency, performance decline, or resource bottlenecks.
Therefore, employing highly efficient training and serving strategies becomes particularly crucial. Many researchers have actively explored and proposed effective methods for training and serving. However, different approaches have different application scenarios. So, it poses a challenge for system developers who struggle to identify the most suitable method for their problems. This challenge is precisely why this paper was proposed.

Although there have been some surveys on foundation models, Most surveys\cite{survey2,survey3,survey4,survey5,survey6,survey7,survey8} predominantly focus on model design and downstream task adaptation, with only a minority delving into foundation model training. However, there are two notable shortcomings in these training-centric surveys\cite{survey1}: firstly, they lack in-depth exploration from the perspective of updates in network, computing, and storage; secondly, their primary emphasis is on the training phase, neglecting considerations for the serving phase.
Therefore, a comprehensive survey of foundation model training and serving methods is essential for system developers and researchers. Accordingly, this paper presents an in-depth analysis of the state-of-the-art methods in this domain.
This paper provides systems developers and researchers valuable information through comprehensive analysis and comparison. It assists them in making the right decisions when confronted with the challenges associated with foundation model systems.
\section{BASIC CONCEPTS}
This section comprehensively explains the fundamental concepts in foundation model systems.
\begin{figure}[t]
  \centering
  \includegraphics[width=.5\textwidth]{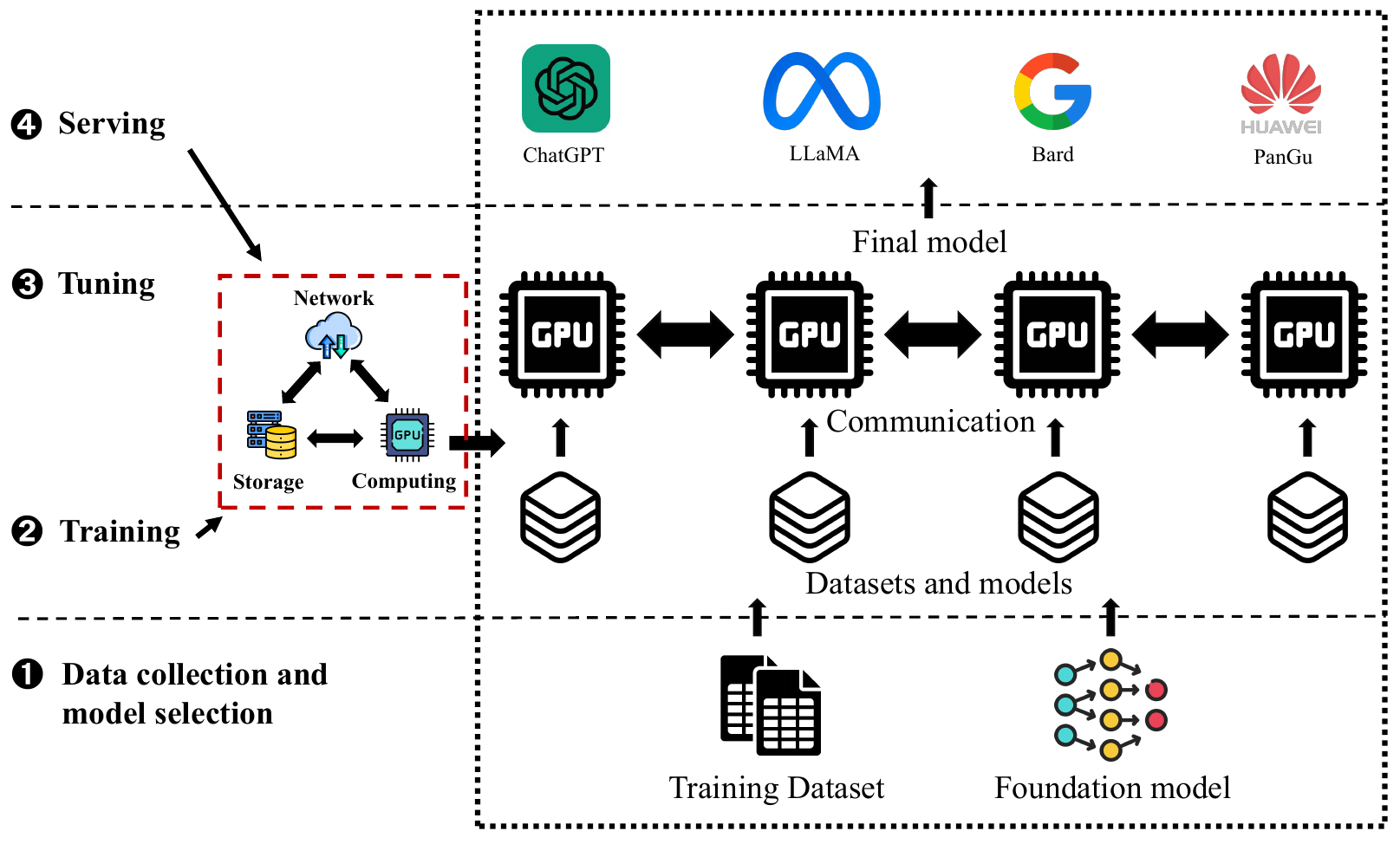}
  \caption{The lifecycle of the foundation model system. } %caption是图片的标题
  \label{lifecycle}   
\end{figure}
\subsection{The lifecycle of the foundation model system}
The lifecycle of the foundation model system (Figure \ref{lifecycle}) encompasses several crucial stages. \ding{202} Initially, the collection and preprocessing of data ensure the quality and availability required for model training. Subsequently, choosing an appropriate model. \ding{203} Transitioning to the training phase, the model undergoes adjustments through the backpropagation algorithm, demanding substantial computational resources to enhance its fitting capability to the training data. \ding{204} Model evaluation and fine-tuning involve assessing performance with test data and adjusting for improved generalization.
Once the model performs satisfactorily, it can be deployed into practical applications. \ding{205} In the serving stage, effective deployment and integration are crucial to ensuring harmonious collaboration with existing systems. The primary focus in this phase centers on performance optimization, aiming to enhance serving speed and reduce latency through strategies such as model quantization and hardware acceleration.
\subsection{Transformer for foundation models}
Transformer\cite{transformer} is a deep learning model architecture comprised of encoders and decoders. Its core innovation lies in the self-attention mechanism, an important component widely utilized in foundational models. The main idea is to enable the model to focus on dynamic associations between different positions, thereby better capturing long-distance interdependent features in a sentence. In the current field of deep learning, the Transformer architecture has become the preferred choice for numerous foundational models. This architecture stands out for its outstanding performance and flexibility, particularly in excelling at natural language processing tasks\cite{bert}. Many pivotal foundational models, such as GPT, LLaMA, and PengCheng Mind, have adopted the design of the Transformer. The successful applications of the Transformer architecture demonstrate its universality in foundational models, providing powerful modeling tools for various tasks.

\begin{table*}[h]
\centering
\caption{Overview of network, computing, and storage optimization strategies in foundation model training.}
\small 
\renewcommand{\arraystretch}{1.0}
\resizebox{1\textwidth}{!}{ % Adjusted the width to 90% of text width
\begin{tabular}{|c|l|c|l|l|c|c|} % Added a new column for Open Resource
\hline
\multicolumn{6}{|c|}{\textbf{Parallel Computing}} \\ \hline
\textbf{Parallelism Type} & \textbf{Specific Strategy} & \textbf{Year} & \textbf{Main Features} & \textbf{Scales (M, B, 10B, 100B+)} & \textbf{Open Resource} \\ \hline
\multirow{3}{*}{Data Parallelism}
 & DDP\cite{DDP} & 2020 & Bucketing gradients, Skipping gradient synchronization & M & \checkmark \\ \cline{2-6}
  & Xu et al. \cite{automatic} & 2020 & Automatic cross-replica sharding & M & $\times$ \\ \cline{2-6}
 & FSDP\cite{FSDP} & 2023 & Fully sharded data parallel & 100B+ & \checkmark \\ \hline
\multirow{4}{*}{Tensor Parallelism} & Megatron-LM \cite{megatron-lm} & 2021 & Weight
matrix partitioned & 100B+ & \checkmark \\ \cline{2-6}
 & Optimus \cite{2d} & 2023 & Scalable 2D-partition paradigm & 100B+ & \checkmark \\ \cline{2-6}
 & Tesseract\cite{2.5d} & 2023 & 2.5-Dimensional Matrix Multiplication & B & $\times$ \\ \cline{2-6}
 & 3D Tensor Parallelism\cite{3d} & 2021 & 3-dimensional
model parallelism, Perfect
load balance & M & $\times$ \\ \hline
\multirow{14}{*}{Pipeline Parallelism} & GPipe\cite{gpipe} & 2019 & Novel batchsplitting pipelining  & 10B & \checkmark \\ \cline{2-6}
 & PipeDream\cite{pipedream} & 2019 & 1F1B, Weight stashing, Vertical sync  & M & $\times$ \\ \cline{2-6}
 & Megatron-LM\cite{megatron-lm} & 2021 & Schedule with Interleaved Stages & 100B+ & \checkmark \\ \cline{2-6}
 & Chimera\cite{chimera} & 2021 & Bidirectional pipelines & B & \checkmark \\ \cline{2-6}
 & PipeDream-2BM\cite{pipedream-2bm} & 2021 & Memory-efficient pipeline parallelism & B & $\times$  \\ \cline{2-6}
    & FTPipe\cite{ftpipe} & 2021 & Mixed-pipe partitioning, Recomputation & 10B & \checkmark \\ \cline{2-6}
  & DAPPLE\cite{dapple} & 2022 & Synchronous training & B & $\times$ \\ \cline{2-6}
   & Varuna\cite{varuna} & 2022 & Recomputation & 100B+ & \checkmark \\ \cline{2-6}
 & Hanayo\cite{hanayo} & 2023 & Wave-like pipeline & 100B+ & $\times$ \\ \cline{2-6}
  & Mixpipe\cite{mixpipe} & 2023 & Mixed scheduling, Flexible bidirectional pipeline & 10B & $\times$ \\ \cline{2-6}
    & Avgpipe\cite{avgpipe} & 2023 & Elastic averaging training method & M & $\times$ \\ \cline{2-6}
& Bpipe\cite{bpipe} & 2023 & Activation balancing & 100B+ & $\times$ \\ \cline{2-6}
& Bamboo\cite{bamboo} & 2023 & Redundant computation, Use of preemptible instances & M & $\times$ \\ \cline{2-6}
 & Dynapipe\cite{dynapipe} & 2024 & Dynamic micro-batching & 10B & \checkmark \\ \hline
\multirow{4}{*}{Expert Parallelism} & GShard\cite{gshard} & 2021 & Sparsely-Gated Mixture-of-Experts & 100B+ & $\times$ \\ \cline{2-6}
 & FastMoE\cite{fastmoe} & 2021 & Open-source
system based on
PyTorch & 100B+ & \checkmark  \\ \cline{2-6}
 & FasterMoE\cite{fastermoe} & 2022 & Dynamic shadowing
method, Novel
roofline-like model & B & \checkmark \\ \cline{2-6}
 & Lina\cite{lina} & 2023 & Tensor partitioning, Two-phase scheduling & B & $\times$ \\ \cline{2-6}
 & Janus\cite{janus} & 2023 & Data-centric paradigm & M & $\times$ \\ \cline{2-6}
 & SmartMoE\cite{smartmoe} & 2023 & Expert placement strateg & 10B & $\times$ \\ \hline
\multirow{3}{*}{Hybrid Parallelism} & Alpa\cite{alpa} & 2022 & Two-level parallel execution, ILP Formulation  & 100B+ & \checkmark \\ \cline{2-6}
 & Smith et al. \cite{dtp} & 2022 &  3D parallelism methodology & 100B+ & $\times$ \\ \cline{2-6}
 & Galvatron\cite{galvatron} & 2022 & Decision tree approach & 10B & \checkmark \\ \hline

 \multicolumn{6}{|c|}{\textbf{GPU Memory Optimization}} \\ 
 \hline
\textbf{Category} & \textbf{Specific Strategy} & \textbf{Year} & \textbf{Main Features}  & \textbf{Scales (M, B, 10B, 100B+)}  & \textbf{Open Resource} \\ \hline
\multirow{2}{*}{Checkpointing and Recomputation } 
& Chen et al.\cite{gradientcheckpoint} & 2016 & Checkpointing,  Recomputation & M& $\times$ \\ \cline{2-6}
& Checkmate\cite{checkmate} & 2022 &  Integer linear program & M& \checkmark \\ \hline

   \multirow{2}{*}{Mixed Precision Training} & Mixed Precision Training\cite{mixed} & 2018 & Mixed precision & M& $\times$ \\ \cline{2-6}
        &  Jia et al.\cite{high} & 2018 &  LARS algorithm & M& $\times$ \\ \hline
\multirow{8}{*}{Memory Swapping} 
        & SwapAdvisor\cite{swapadvisor} & 2020 &Enetic algorithms &M& $\times$ \\ \cline{2-6}
        & Autotm\cite{autoTM} & 2020 & Integer linear programming& M& \checkmark \\ \cline{2-6}
        & FlashNeuron\cite{flashneuron} & 2021 & SSDs for data offloading and
prefetching & B& \checkmark \\ \cline{2-6}
        & Stronghold\cite{stronghold} &  2022 & Work window method &10B & \checkmark \\ \cline{2-6}
        & Patrickstar\cite{patrickstar} & 2022 & Chunks &10B & \checkmark \\ \cline{2-6}
        & G10\cite{g10} & 2023 & Amalgamating GPU,
host, flash memory into a unified memory
space & B & \checkmark \\ \cline{2-6}
        & DeepUM\cite{deepum} & 2023 & Enhances Unified Memory
by prefetching techniques &B& $\times$ \\ \hline
\multirow{3}{*}{Zero Redundancy Optimization} & ZeRO\cite{zero} & 2020 & Zero Redundancy &100B+&\checkmark \\ \cline{2-6}
        & ZeRO-Offload\cite{zero-offload} & 2021 &Parameters is offloaded to CPU memory &10B& \checkmark \\ \cline{2-6}
        & ZeRO-Infinity\cite{zero-infinity} & 2021 & Parameters is offloaded to CPU, and
NVMe memory &100B+& \checkmark \\ \hline
    \multicolumn{6}{|c|}{\textbf{Communication Optimization}} \\ \hline
      \textbf{Category} & \textbf{Specific Strategy} & \textbf{Year} & \textbf{Main Features}  & \textbf{Scales (M, B, 10B, 100B+)}  & \textbf{Open Resource} \\ \hline
        \multirow{4}{*}{Communication Optimization}
        & Bagua\cite{bagua} & 2021 & MPI-style communication library & M & \checkmark \\ \cline{2-6}
        & Out-Of-Order BackProp\cite{outoforder} & 2022 & Out-Of-Order Computation &100B+ & \checkmark \\ \cline{2-6}
        & ZeRO++\cite{zero++} & 2023 & Weight quantization &100B+&$\checkmark$ \\ \cline{2-6}
        & Wang et al.\cite{overlap} & 2023& Decomposing the original communication and computational
operations & 100B+& $\times$ \\ \cline{2-6}
        & Mobius\cite{mobius} & 2023 & Cross-mapping
strategy & B & $\times$ \\ \cline{2-6}
        & Optimus-CC\cite{optimus-cc} & 2023 & Compression& B & $\times$ \\ \cline{2-6}
        & COCKTAILSGD \cite{COCKTAILSGD} & 2023 & Communication compression &10B& $\times$ \\ \hline
\end{tabular}}
\label{tab:parallel-strategies}
\end{table*}

\section{MODEL TRAINING}
\begin{figure*}[t]
  \centering
  \includegraphics[width=1\textwidth]{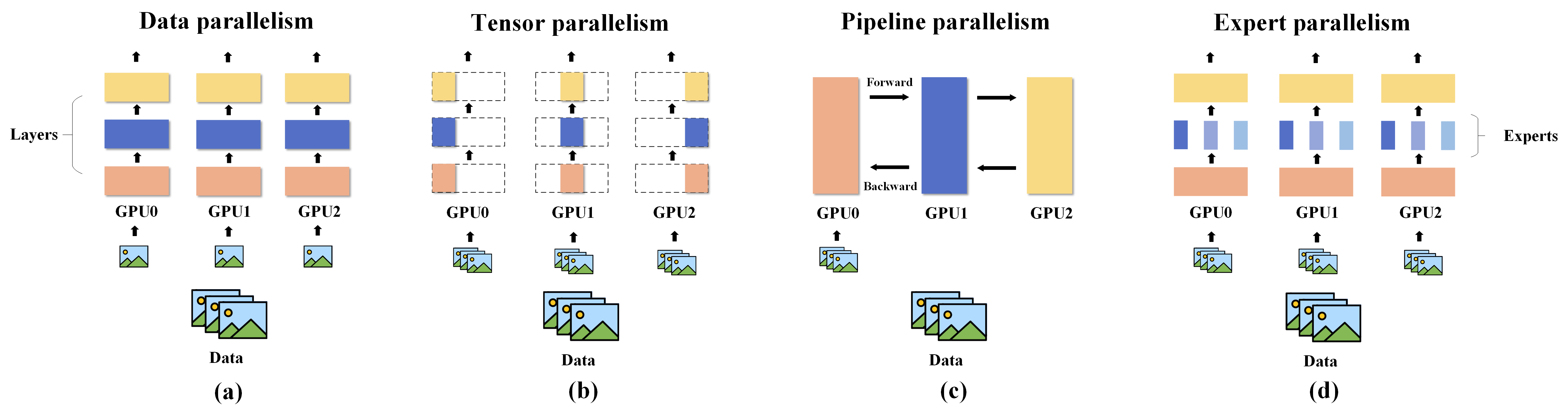}
  \caption{Schematic diagram of parallelization strategies in foundation model systems. Different color blocks indicate different layers in the network.} %caption是图片的标题
  \label{parallel}
\end{figure*}
In foundation model training, the most significant challenges are the high demands for memory and computational power. Therefore, this section explores the implementation of optimization strategies in foundation model training from three 
perspectives, network, computing, and storage, to address these challenges, as shown in Table \ref{tab:parallel-strategies}.

\subsection{Advanced Techniques in Parallel Computing}
\subsubsection{Data Parallelism: Accelerating Workloads Effectively}
In data parallelism, each computational node possesses a replica of the model and independently processes a subset of data assigned to it. As shown in Figure \ref{parallel}a, each node uses its model replica for forward and backward propagation and gradient calculation. So, it requires gradient aggregation and synchronization operations to update the global model parameters. This distributed approach significantly reduces the computational load on individual nodes and speeds up the training process by parallelizing the workload.

Distributed Data Parallel (DDP)\cite{DDP} utilizes gradient bucketing, computation-communication overlap, and gradient synchronization skipping to enhance the efficiency of distributed data parallelism. The Ring-AllReduce method\cite{sjbxring} effectively addresses the communication load issue in data parallelism with its unique design. In this method, worker nodes are organized in a ring topology, communicating only with adjacent nodes. 
In the aforementioned data parallelism approach, storing the entire model's parameters on each node simplifies training but significantly increases memory demand, especially for foundation models. To solve this problem, several solutions have been proposed. Facebook introduced a technique called Fully Sharded Data Parallel (FSDP)\cite{FSDP} to tackle this issue. It divides model parameters into smaller units, restoring the complete model parameters through communication before computation and discarding them immediately after the calculation. Similarly, Xu et al.\cite{automatic} proposed an automatic cross-replica sharding technique for weight updates in data-parallel training to optimize memory and communication efficiency during model training.

\subsubsection{Tensor Parallelism: Scaling Foundation Models}
Tensor parallelism is developed to address the challenges of training foundation models that exceed the memory capacity of a single device. In tensor parallelism (As
shown in Figure \ref{parallel}b), the parameters and computations of the model are divided (or sliced) across multiple computing devices, effectively reducing the memory load on each device.

Megatron-LM\cite{megatron-lm} introduced an efficient form of 1D tensor parallelism. 
For a given computational task, it involves two GEMM operations and a GeLU non-linearity:
$Y=\operatorname{GeLU}(X A), \quad Z=Y B$. $A$ can be partitioned into $\left[\begin{array}{ll}A_1,A_2\end{array}\right]$. So each processor can independently compute the $Y_i$: $$
\left[Y_{1}, Y_{2}\right]=\left[\operatorname{GeLU}\left(X A_{1}\right), \operatorname{GeLU}\left(X A_{2}\right)\right].
$$ The second weight matrix $B$ can be split into $\left[\begin{array}{l}B_1 \\ B_2\end{array}\right]$, so $Z$ is equal to $\left[\begin{array}{ll}Y_1,Y_2\end{array}\right]\left[\begin{array}{l}B_1 \\ B_2\end{array}\right].$
With this approach, $Y_i B_i$ can be computed separately on individual processors. 
In Transformer models, the method of 1D tensor parallelism is effectively applied to the computation of multi-head attention. This allows multiple processing units to simultaneously calculate different attention heads without waiting for the results of others. 

Optimus\cite{2d} proposed an efficient and scalable 2D tensor parallelism. 
It is introduced based on the scalable universal matrix multiplication algorithm (SUMMA)\cite{summa}. Compared to 1D tensor parallelism, 2D parallelism distributes the computational load across more processing units, significantly enhancing overall computational efficiency.
Although 2D parallelism offers a more fine-grained model partitioning approach, it can introduce higher communication overhead. To solve this, 2.5D tensor parallelism\cite{2.5d} is introduced, building upon 2.5D matrix multiplication\cite{2.5dm}, which leverages additional devices to minimize communication requirements. 

To balance computation, memory, and communication loads effectively, 3D tensor parallelism\cite{3d} employs a 3-D Parallel Matrix Multiplication algorithm to accurately map and execute the computation process of the Transformer model.
This algorithm optimizes the use of computational resources by intelligently distributing and computing different parts of the input and weight matrices on designated processors.

\subsubsection{Pipeline Parallelism: Enhancing Foundation Model Scalability}
In pipeline parallelism (Figure \ref{parallel}c), the entire model is divided into several stages, with each part allocated to an independent GPU. However, a typical issue in pipeline parallel processing is the idle time created due to waiting for dependent data or processing results, commonly referred to as the bubble phenomenon. Therefore, effectively reducing these bubbles to enhance GPU utilization in pipeline parallelism becomes a critical issue.

\begin{figure}[t]
  \centering
  \includegraphics[width=.5\textwidth]{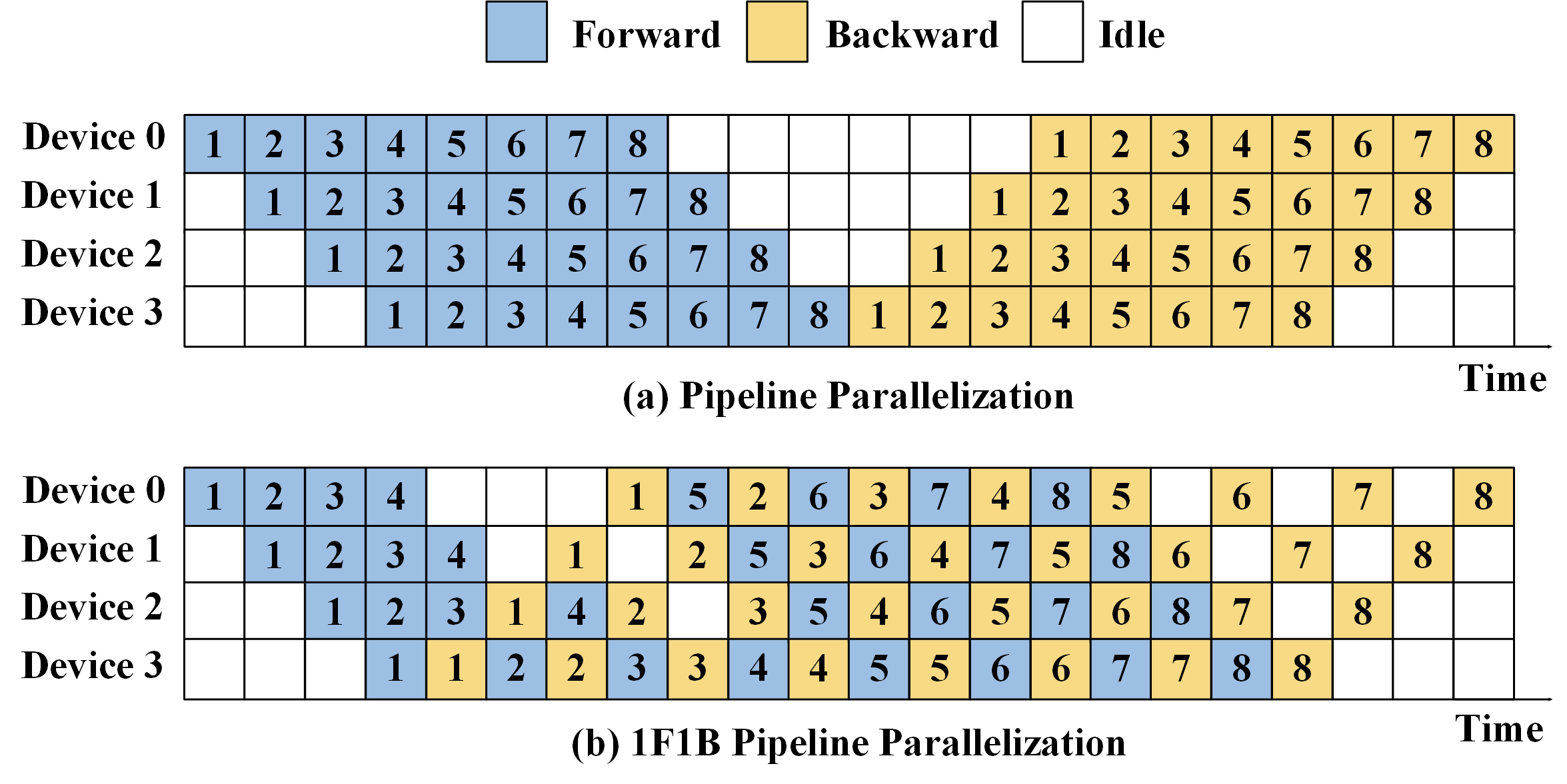} %1.png是图片文件的相对路径
  \caption{Schematic diagram of pipeline parallelization and 1F1B pipeline parallelization.} %caption是图片的标题
  \label{img} %此处的label相当于一个图片的专属标志，目的是方便上下文的引用
\end{figure}

GPipe\cite{gpipe} is one of the first significant works to apply the concept of pipeline parallelism to the training of foundation models.
However, GPipe requires waiting for each micro-batch to complete forward propagation before starting backward propagation, as shown in Figure \ref{img}a. Therefore, the intermediate results (activations) produced during the forward computation of each micro-batch need to be cached in memory for subsequent backpropagation, resulting in increased memory usage. Meanwhile, this approach can also lead to the creation of a significant number of bubbles.

So PipeDream\cite{pipedream} utilizes a one-forward-one-backward (1F1B) strategy, as shown in Figure \ref{img}b, to solve these 
problems, in which the backward propagation process immediately follows the completion of forward propagation for a micro-batch.
However, this training mode introduces two types of parameter inconsistency. PipeDream utilizes weight stashing and vertical sync methods to address these issues. Varuna\cite{varuna} improves upon PipeDream by performing recomputation earlier during the backward pass, effectively reducing bubbles and memory usage. Meanwhile, similar to PipeDream, DAPPLE\cite{dapple} performs synchronization after completing the forward and backward propagation for micro-batches. This synchronization ensures the consistency of model parameters across micro-batches, avoiding parameter inconsistency issues.

PipeDream utilizes a weight storage scheme to use the same weight version in forward and backward propagation for the same input. In the worst case, the number of stored weight versions equals the pipeline depth. Therefore, this can result in increased memory consumption. PipeDream-2BM\cite{pipedream-2bm} maintains only two versions of model weights within the pipeline. By storing only two versions, it significantly reduces memory usage. 
In Megatron-LM\cite{megatron-lm}, pipeline parallelism is implemented using an Interleaved Schedule approach. To reduce pipeline bubbles, the Interleaved Schedule method assigns each device to two sets of model chunks, allowing each device to handle multiple stages. By utilizing the idle time of devices that would otherwise be waiting for backward computation, it can perform forward computation for the second set of model chunks, effectively reducing the size of bubbles in the pipeline.
In contrast, Chimera\cite{chimera} utilizes a bidirectional pipeline by deploying multiple stages of the model on a single GPU. Chimera minimizes idle time and maximizes GPU utilization by interleaving the computation of forward and backward propagation across different stages. Different from Chimera, Hanayo\cite{hanayo} avoids the strategy of model replication. Instead, it employed a wave-like pipeline scheme, further reducing bubble rates and enhancing performance. Similarly employing a bidirectional pipeline, MixPipe\cite{mixpipe} achieves a better balance between pipeline and device utilization by adjusting the number of micro-batches. Additionally, MixPipe designs a hybrid scheduling strategy that combines 1F1B and 2F1B to achieve a more balanced memory usage, further decreasing bubble rates. To further reduce bubbles, AvgPipe\cite{avgpipe} employs the approach of multiple parallel pipelines, with each pipeline handling a batch of data in each iteration. It trains parallel models using an elastic averaging training method. By processing more batches, 
AvgPipe can subdivide each batch into finer-grained micro-batches, effectively reducing the generation of bubbles.

In traditional pipelines, feeding a batch of samples with equal lengths into the GPU for training is common. In each batch, padding is applied to the input sequences to accommodate the length of the longest sequence, leading to evident memory wastage. Dynapipe\cite{dynapipe} introduces a method of dynamic micro-batching, where the core idea is to ensure consistent sequence lengths among samples within each micro-batch without requiring uniformity across micro-batches. This approach reduces the padding overhead in micro-batch processing, effectively lowering memory consumption. To address memory balancing in the pipeline, Bpipe\cite{bpipe} uses an activation balancing approach that ensures that all GPUs can fully utilize comparable amounts of memory by transferring intermediate activations between different GPUs during training. This innovation solves the problem that some GPUs may face high memory pressure while others fail to fully utilize the performance.

The continuous growth of the foundation's scale has triggered a substantial demand for training resources. In addressing this issue, Bamboo\cite{bamboo} significantly reduces training costs by using preemptible instances optimally. When idle, these instances are available at a lower cost but may be preempted once users submit priority requests. Bamboo optimizes the training pipeline by introducing redundant computations to overcome this challenge. Specifically, each node performs computations not only on its layer but also on adjacent layers. Bamboo cleverly incorporates these additional computations into redundant layers, thus providing greater flexibility at a lower cost.
The pipeline methods previously discussed primarily involve a simplistic partitioning of a model's adjacent layers. This approach can lead to imbalanced workload distribution across GPUs. As an improvement, FTPipe\cite{ftpipe} introduces mixed-pipe partitioning technology. It employs a heuristic algorithm to allocate GPUs based on any computational blocks in the computation graph, not just adjacent layers.

\subsubsection{Expert Parallelism: Enhancing Specialized Computing Capabilities}

Expert parallelism, as depicted in Figure \ref{parallel}d, involves segmenting a specific part of a model into several specialized sub-models, referred to as experts, and distributing them across various computational devices. A gating network is used to determine how to allocate input data efficiently among these different experts.

Google's GShard\cite{gshard} introduces the MoE\cite{moe} structure for the first time in training foundation Transformer-based models, aiming to address scalability issues in foundation model training. To optimize the performance of the MoE model, FastMoE\cite{fastmoe} was proposed. It is the first high-performance MoE open-source system that supports the PyTorch\cite{pytorch} framework. FastMoE pairs the FFN layer in the Transformer and adopts a finer parallelization strategy. This strategy significantly speeds up the computation of the FFN part of the Transformer model.
During the training process of MoE systems, challenges such as dynamic load imbalance and congested end-to-end communication need to be addressed. To tackle these challenges, FasterMoE\cite{fastermoe} proposes a dynamic shadowing method to handle load imbalances. By dynamically adjusting task allocation and scheduling, system resources are utilized more evenly, improving overall efficiency.

On the other hand, SmartMoE\cite{smartmoe} system provides comprehensive support for distributed training strategies. 
To address the dynamic computational workload of MoE models, the SmartMoE system introduces a unique expert placement strategy. Building upon the classic combination of parallel strategies, this strategy achieves dynamic load balancing. By intelligently adjusting the deployment positions of various experts in the model, the SmartMoE system effectively balances the computational workload and improves overall system efficiency.  In distributed data parallelism, contention may occur between the all-to-all communication among MoEs and the all-reduce operations, leading to prolonged training times. Therefore, Lina\cite{lina} integrates tensor partitioning and pipelining to perform micro-operation scheduling, reducing blocking periods in distributed training. All of the above approaches use an expert-centric paradigm, keeping the expert in place and providing information to the expert through an all-to-all process. However, Janus\cite{janus} proposes a new data-centric paradigm: maintaining the data in place and moving the experts between GPUs. Janus hides the communication time by scheduling the requests of fetching experts in a fine-grained manner, thus reducing cross-node traffic. Moreover, Janus develops a topology-aware priority strategy, ensuring smooth intra-node expert exchanges without resource contention.

\subsubsection{Hybrid Parallelism: Combining the Power of Different Parallel Computing Approaches}
Although various parallel technologies have shown significant effects in theoretical and experimental research, a single parallel strategy often fails to meet the growing computational demands and complexity in actual deep learning model training. Therefore, hybrid parallelism becomes critical to addressing this challenge. 
The core of hybrid parallelism lies in its ability to make customized strategy choices based on the specific requirements of the task and available hardware resources, thereby maximizing training efficiency while ensuring model performance.

Combining multiple parallelization techniques for enhanced efficiency is common when conducting pre-training of foundation models with parameter scales in tens to hundreds of billions. Smith et al.\cite{dtp} utilized a combination of pipeline and tensor parallelism techniques to parallelize the Transformer block in Megatron-Turing NLG during their training using DeepSpeed\cite{deepspeed} and Megatron-LM. They expanded the training scale by incorporating data parallelism, allowing for training on more GPUs.
To simplify the application and enhance the efficiency of parallelization strategies, Alpa\cite{alpa} integrates all parallelization strategies into a single framework, establishing a compiler that automatically generates optimal parallelization strategies. Similarly, Galvatron\cite{galvatron} introduces a decision tree approach that leverages logical intuition for pruning, thereby significantly cutting down the search space. In addition, Galvatron employs a dynamic programming search algorithm to determine the most effective hybrid parallelization strategy.

\subsection{GPU Memory Optimization In Training}
As the model size increases, the demand for GPU memory grows exponentially. However, limited by hardware resources, insufficient GPU memory often becomes a bottleneck, restricting the scale and performance of the training. Therefore, developing effective GPU memory optimization techniques is essential to reduce memory consumption. Subsequent sections will explore various innovative GPU memory optimization techniques targeting these overheads.

\subsubsection{Checkpointing and Recomputation for Memory Efficiency}
In foundation model training, activation checkpointing technology reduces memory consumption by only saving key activation values and uses recomputation technology to regenerate these values during backpropagation. This combined approach effectively balances memory usage and computational efficiency, enabling the training of foundation models even with limited hardware resources.

It was Chen et al.\cite{gradientcheckpoint} who first proposed the concept of activation checkpointing to tackle the high memory consumption in foundation model training. By selectively removing unneeded intermediate activations in the forward propagation process and reconstructing them during backward propagation through additional computations, this method significantly reduces GPU memory usage while allowing for the training of more extensive networks.
However, recomputation imposes an additional time overhead. Therefore, it requires a trade-off between training time and memory requirements. To address this problem, Jain et al. proposed the Checkmate\cite{checkmate}, which models the problem to minimize computation time while ensuring that task scheduling does not exceed the memory limit of the device.
The Checkmate effectively manages memory usage by dynamically determining when to store activations and recompute them. This enables the training of larger-scale networks within the constraints of limited memory resources, providing an effective solution to address memory limitations in foundation model training.

\subsubsection{Optimizing with Mixed Precision Training}
Mixed Precision Training\cite{mixed} is a technique used in foundation models that simultaneously employs both low-precision and high-precision data types. Representing the training data types as 16-bit floating-point numbers can reduce the amount of computation while lowering the memory requirement. However, the 16-bit floating-point representation will inevitably impact model convergence. Jia et al.\cite{high} utilized the LARS algorithm\cite{image} to solve this problem. The algorithm works by using different learning rates for different layers. However, the test found that applying the LARS algorithm to the training of half-precision models directly caused a great loss of accuracy. This is because after multiplying by the LARS coefficients, many parameters directly go to zero due to the small range of the half-precision values, so Jia et al. converted the half-precision parameters into single-precision and then combined them with the LARS.
\subsubsection{Memory Swapping Techniques in Optimization}
The basic idea of memory swapping technology is to offload the computational burden from the GPU to other devices, such as CPUs or NVMe. It migrates some model parameters and computational tasks from the GPU to other devices. This relieves the GPU's workload and enables it to handle the remaining computational tasks more efficiently.

This idea was first introduced in vDNN\cite{vdnn}, which aims to reduce the pressure on the GPU memory by moving data that does not require immediate access from the GPU to the CPU memory. 
The implementation of vDNN represents an initial application of swapping technology. However, with technological advancements, more sophisticated methods have emerged. SwapAdvisor\cite {swapadvisor} employs genetic algorithms to automatically search for the best data transfer strategy as an alternative to manual judgment-based approaches. The benefit of this automated approach is that it reduces the need for human intervention, thereby increasing efficiency. In contrast, Autotm\cite{autoTM} uses an integer linear programming approach to search for suitable transfer strategies.

Stronghold\cite{stronghold} introduces a work window method, which keeps only part of the model's layers and parameters in the GPU. Under this mechanism, the GPU processes only the model layers within the work window, transferring the rest to the CPU. The corresponding resources are only moved from the CPU to the GPU when the work window shifts. Additionally, Stronghold models the window size, and leverages computation and communication overlap to hide the communication costs between the CPU and GPU effectively.
Meanwhile, FlashNeuron\cite{flashneuron} considers that offloading data directly to the CPU might interfere with other tasks running on the CPU and thus uses SSDs for data offloading and prefetching. DeepUM\cite{deepum} enhances Unified Memory (UM) by incorporating prefetching techniques, effectively reducing the additional overhead caused by address translations and page faults. Similarly, G10\cite{g10} innovatively extends the Unified Memory of GPUs, amalgamating GPU memory, host memory, and flash memory into a unified memory space. This fusion is achieved by storing flash memory page addresses in the UM page table. Consequently, a unified page table can point to host, GPU, or flash memory addresses. By preemptively analyzing the lifecycle of tensors, G10 enables efficient tensor swapping when needed, maximizing the overlap between GPU computation and tensor migration.
Furthermore, Patrickstar\cite{patrickstar} proposes a memory management method based on chunks, a series of consecutive tensors of the same size. 
This method is similar to storing files in fixed-sized disk blocks in a distributed file system. During training, chunks with different lifecycles can share memory, reducing memory usage. Additionally, Patrickstar collects memory usage information during the warm-up iteration phase to optimize memory management.
% \begin{table}[h]
% \centering
% \caption{Overview of Communication Optimization Techniques in Large-Scale Model Training}
% \small % Reduce font size
% \renewcommand{\arraystretch}{1.0} % Reduce row spacing
% \begin{tabular}{|m{2.5cm}|c|c|} % Adjusted column widths
% \hline
% \textbf{Technique} & \textbf{Year} & \textbf{Publication} \\ \hline
% ZeRO++\cite{zero++} & 2023 & arXiv \\ \hline
% Shibo Wang et al. Method\cite{overlap} & 2023 & ASPLOS \\ \hline
% Mobius\cite{mobius} & 2023 & ASPLOS \\ \hline
% Optimus-CC\cite{optimus-cc} & 2023 & ASPLOS \\ \hline
% \end{tabular}
% \label{tab:communication-optimization-structured}
% \end{table}

Beyond the methods above, other works like ZeRO-Offload and ZeRO-Infinity have also employed Memory Swapping Techniques. To comprehensively introduce the ZeRO series of research, this paper includes these additional works in the next section.
\subsubsection{Zero Redundancy Optimizers}
Microsoft has developed a technology called Zero Redundancy Optimization (ZeRO)\cite{zero} as the core of the DeepSpeed distributed training framework. The core idea of ZeRO is to reduce the GPU memory by sacrificing some of the communication overhead. ZeRO divides the model parameters, gradients, and optimizer states into multiple parts, with each GPU maintaining only a portion of them during training and obtaining the rest when needed through an AllGather operation.
Building upon the foundation laid by ZeRO, ZeRO-Offload\cite{zero-offload} leverages the idea of Heterogeneous DL training\cite{dl-training2} to alleviate the pressure on GPU memory by effectively utilizing CPU memory. It divides the model parameters into two parts. One part of the parameters is kept in GPU memory for efficient computation during forward and backward propagation. The other part of the parameters is offloaded to CPU memory and accessed when needed.
Further advancing these concepts, ZeRO-Infinity\cite{zero-infinity}, similar to ZeRO-Offload, leverages GPU, CPU, and NVMe memory to enable the training of foundation models on limited resources without the need for code refactoring. With ZeRO-Infinity, the model parameters and gradients are still computed on the GPU, while the optimizer state and activations are offloaded to more suitable NVMe memory and CPU, respectively.

\subsection{Communication Optimization}
As demonstrated in the previous sections, communication overhead is a significant bottleneck in the distributed training of foundation deep learning models. This issue is especially pronounced when synchronizing model parameters, gradients, and optimizer states across multiple GPUs or nodes. The mainstream solutions focus on reducing the amount of communication, optimizing communication patterns, and enhancing the overlap between computation and communication.

For instance, Gan et al. developed an MPI-style communication library called Bagua\cite{bagua}. The library provides a series of flexible and modular primitives to support state-of-the-art system relaxation techniques of distributed training. Bagua achieves efficient implementation and scalability through this design for various cutting-edge distributed learning algorithms.
The method proposed by Wang et al.\cite{overlap} involves decomposing the original communication and computational operations into more fine-grained tasks, thereby achieving an overlap between communication and computation that effectively reduces data communication overhead. 
Mobius\cite{mobius} introduces a pipeline strategy for heterogeneous memory, which overlaps communication with computation by prefetching data from the CPU to the GPU memory for the next stage. Additionally, it employs a Cross-mapping strategy to reduce communication contention, further optimizing overall performance.
Simultaneously, Out-Of-Order BackProp\cite{outoforder} maximizes the overlap between communication and computation by optimizing the sequence of computing output gradients, weight gradients, and parameter updates.

ZeRO achieves parallel computation by distributing model weights, gradients, and optimizer states across multiple GPUs, increasing communication volume and frequency. As an improvement, ZeRO++\cite{zero++} employs weight quantization, meaning model parameters are compressed into smaller data types (such as INT8) in real-time before communication, reducing the required communication bandwidth and time. Moreover, ZeRO++ maintains a complete model copy on each machine, enhancing intra-machine communication bandwidth.
COCKTAILSGD\cite{COCKTAILSGD} integrates various communication compression techniques, cleverly overlapping communication with local gradient computation. During the communication steps, it combines three different compression techniques (random sparsification, top-K sparsification, and quantization) to achieve more excellent compression than each method individually.
Lastly, Optimus-CC\cite{optimus-cc} utilizes three techniques: compression of back-propagation gradients, merging of embedding layer synchronization operations, and selective phase compression to reduce inter-node communication volume. Optimus-CC selectively compresses based on the communication needs of different training stages, thus minimizing unnecessary communication overhead and enhancing overall training efficiency.

\section{MODEL SERVING}
This section discusses five principal areas of optimization in foundation model serving systems: batch processing optimization, sparse acceleration techniques, resource scheduling optimization, GPU memory optimization, and multi-model inference (As shown in Table \ref{tab:inference}). It presents various innovative techniques and technologies designed to enhance processing efficiency, minimize latency, and improve memory usage. These strategies are categorized within the ``network-computing-storage" optimization framework.

\begin{itemize}
\item \textbf{Network optimization} is accomplished through efficient batch processing and resource scheduling, optimizing data flow and task execution.
\item \textbf{Computing optimization} is characterized by multi-model inference, enabling the efficient utilization of computational resources.
\item \textbf{Storage optimization} involves GPU memory management and the application of sparse acceleration techniques, collectively reducing memory footprint and computational overhead.
\end{itemize}

Integrating these ``network-computing-storage" principles ensures a comprehensive optimization approach, which is crucial for the performance of foundation model serving systems.

\begin{table}[]
\centering
\caption{Summary of Optimization Techniques in Foundation Model Serving}
\begin{tabular}{|lll|}
\hline
\multicolumn{1}{|l|}{Method/Framework}                                                & \multicolumn{1}{l|}{Main Features}                  & Year \\ \hline
\multicolumn{3}{|l|}{\textit{\textbf{Batch Processing Optimization}}}                                                                              \\ \hline
\multicolumn{1}{|l|}{DVABatch\cite{dvabatch}}                        & \multicolumn{1}{l|}{Dynamic Batching}               & 2022 \\ \hline
\multicolumn{1}{|l|}{Orca\cite{orca}}                                & \multicolumn{1}{l|}{Selective Batching}             & 2022 \\ \hline
\multicolumn{3}{|l|}{\textit{\textbf{Sparse Acceleration Techniques}}}                                                                             \\ \hline
\multicolumn{1}{|l|}{SparseAttention\cite{sparseAttn}}               & \multicolumn{1}{l|}{Structured sparse attention masks}               & 2023 \\ \hline
\multicolumn{1}{|l|}{Deja Vu\cite{dejavu}}                           & \multicolumn{1}{l|}{Use MLP to predict sparsity}               & 2023 \\ \hline
\multicolumn{1}{|l|}{H2O\cite{h2o}}                                  & \multicolumn{1}{l|}{Sparse KV Cache}                & 2023 \\ \hline
\multicolumn{1}{|l|}{STI\cite{sti}}                                  & \multicolumn{1}{l|}{Sharded Models}                 & 2023 \\ \hline
\multicolumn{1}{|l|}{OliVe\cite{olive}}                              & \multicolumn{1}{l|}{Outlier Quantization}           & 2023 \\ \hline
\multicolumn{3}{|l|}{\textit{\textbf{Resource Scheduling Optimization}}}                                                                           \\ \hline
\multicolumn{1}{|l|}{Clockwork\cite{clockwork}}                      & \multicolumn{1}{l|}{Latency Predict}                & 2020 \\ \hline
\multicolumn{1}{|l|}{DeepSpeed Inference\cite{deepspeedinference}}   & \multicolumn{1}{l|}{Integrated Scheduling}          & 2022 \\ \hline
\multicolumn{1}{|l|}{REEF\cite{reef}}                                & \multicolumn{1}{l|}{Real-Time Preemptive Schedule}  & 2022 \\ \hline
\multicolumn{1}{|l|}{AlphaServe\cite{alphaserve}}                    & \multicolumn{1}{l|}{Automated Model Parallelism}    & 2023 \\ \hline
\multicolumn{1}{|l|}{FastServe\cite{fastserve}}                      & \multicolumn{1}{l|}{Preemptive Scheduling}          & 2023 \\ \hline
\multicolumn{1}{|l|}{SHEPHERD\cite{SHEPHERD}}                        & \multicolumn{1}{l|}{Predictive Load Management}     & 2023 \\ \hline
\multicolumn{1}{|l|}{OSML\cite{osml}}                                & \multicolumn{1}{l|}{Predictive Resource Allocation} & 2023 \\ \hline
\multicolumn{3}{|l|}{\textit{\textbf{GPU Memory Optimization In Inference}}}                                                                       \\ \hline
\multicolumn{1}{|l|}{Gpulet\cite{gpulet}}                            & \multicolumn{1}{l|}{Virtual GPU Partitioning}       & 2022 \\ \hline
\multicolumn{1}{|l|}{FlexGen\cite{flexgen}}                          & \multicolumn{1}{l|}{Zig-Zag Block Scheduling}       & 2023 \\ \hline
\multicolumn{1}{|l|}{DHA\cite{dha}}                               & \multicolumn{1}{l|}{Direct GPU Access}              & 2023 \\ \hline
\multicolumn{1}{|l|}{vLLM\cite{vllm}}                                & \multicolumn{1}{l|}{PageAttention Mechanism}        & 2023 \\ \hline
\multicolumn{3}{|l|}{\textit{\textbf{Multi-Model Inference}}}                                                                                       \\ \hline
\multicolumn{1}{|l|}{PetS\cite{pets}}                                & \multicolumn{1}{l|}{Selective Model Sharing}        & 2022 \\ \hline
\multicolumn{1}{|l|}{Tabi\cite{tabi}}                                & \multicolumn{1}{l|}{Multi-Level Model Inference}    & 2023 \\ \hline
\multicolumn{1}{|l|}{Speculative Decoding\cite{speculativedecoding}} & \multicolumn{1}{l|}{Speculative decoding} & 2023 \\ \hline
\multicolumn{1}{|l|}{LLMCad\cite{llmcad}}                            & \multicolumn{1}{l|}{Model collaboration} & 2023 \\ \hline
\end{tabular}
\label{tab:inference}
\end{table}

\subsection{Batch Processing Optimization}
Batch processing allows models to handle multiple requests efficiently by grouping input data into batches. This method allows for more efficient use of computational resources by leveraging parallel processing capabilities, significantly improving the throughput and reducing the latency of model inferences. DVABatch\cite{dvabatch} proposes a multi-entry and multi-exit strategy that employs operations like \textit{new}, \textit{split}, and \textit{stretch} to dynamically customize batch sizes for different stages of the model, thereby optimizing efficiency, throughput, and reducing latency. In addition to this approach, Orca\cite{orca} introduces a selective batching mechanism that strategically applies batch processing and padding to fully connected layers, maximizing efficiency. Simultaneously, it refrains from applying this method to attention layers, minimizing memory overhead. Furthermore, Orca presents an iterative-level scheduling strategy that offers adaptability by enabling batch size adjustments after each processing iteration.  

\subsection{Sparse Acceleration Techniques}
Sparse acceleration techniques play a crucial role in optimizing the performance of Transformer-based foundation models when faced with limited computational and memory resources. These methods leverage the inherent sparsity in model parameters, attention mechanisms, and KV Cache to prioritize computation and storage. Transformers leverage a KV Cache within the attention mechanism to remember crucial parts of the data, which is essential for efficiency, particularly in autoregressive models that generate new tokens iteratively and would otherwise require costly re-computation of keys and values for each token. By focusing on the most influential components of the model and reducing the overhead on less critical areas, sparse acceleration approaches enable high model performance while facilitating deployment on edge devices and mobile platforms where resources are scarce.

Addressing the computational intensity of self-attention mechanisms in Transformers, Dai et al.\cite{sparseAttn} introduce a method that leverages the intrinsic sparsity within self-attention matrices. The method employs structured sparse attention masks to allocate computational resources to the most influential attention parameters, which are identified by analyzing the attention distribution of the model. These parameters shape the initial mask, designed to conform to specific patterns like blocks or stripes. Through entropy-aware fine-tuning, this mask undergoes further refinement by integrating the entropy of the attention matrix's rows into the model's loss function. Another notable work is Deja Vu\cite{dejavu}, which offers a strategic solution by leveraging the concept of contextual sparsity. The proposed approach effectively identifies and activates selective attention heads and FFN parameters that are crucial in processing the given input. The Deja Vu utilizes an MLP to predict the critical attention heads and FFN parameters precisely. In cases where the KV Cache retrieval fails, it triggers a recomputation process. Another relevant work is that the H2O\cite{h2o} method suggests that in attention blocks, the cumulative attention scores of tokens follow a power-law distribution, with only a few tokens playing a pivotal role in generation. To conserve memory, H2O retains only the KV Cache for these pivotal tokens, which are identified as highly contributive KV Cache based on their elevated cumulative attention scores. This approach substantially reduces the memory requirement while retaining the essential KV Cache of the attention mechanism.

In low-resource scenarios on edge devices, both computation and memory are constrained. STI\cite{sti} tackles memory constraints and I/O delays for foundation models on edge devices by partitioning the model into manageable shards. A central scheduler orchestrates the I/O operations of these shards, considering resource availability and the significance of each shard. It stores shards at multiple precision levels and dynamically selects the optimal precision for each shard based on its importance, thereby optimizing both accuracy and latency. In the realm of model quantization acceleration, one prominent contribution is OliVe\cite{olive}. This study presents the concept of Outlier-Victim Pair (OVP) quantization as a key technique for effectively managing critical outliers with minimal reduction in model accuracy. OliVe's innovative insight lies in recognizing the importance of outliers for model accuracy while identifying adjacent normal values, termed `victim values', that can be pruned without significant performance loss. OliVe strategically retains outliers and prunes adjacent normal values. This selective pruning approach aligns with hardware design principles.

\subsection{Resource Scheduling Optimization}
Effective resource scheduling is essential in optimizing service delivery. DeepSpeed Inference\cite{deepspeedinference} offers a multi-GPU inference solution designed to handle foundation models while adhering to limited GPU memory constraints. It introduces an inference pipeline-parallel schedule specifically tailored for the autoregressive Decoder in generative models, optimizing prompt processing and token generation to minimize latency. Moreover, DeepSpeed Inference leverages both GPU and CPU resources, including NVMe storage, to alleviate the burden on GPU memory. Furthermore, DeepSpeed Inference incorporates operator fusion within Transformer modules, efficiently reducing latency and increasing throughput. In the field of model parallelism research, AlphaServe\cite{alphaserve} enhances foundation model inference by employing model parallelism, which distributes models across multiple GPUs to overcome the limitations of single GPU memory and reduce request latency. AlphaServe incorporates a two-layer placement algorithm that optimizes the distribution of model ensembles in clusters, ensuring compliance with Service Level Objective (SLO) requirements. Building upon the Alpha framework, it automates model parallelism in inference to simplify the management of parallelism. Taking optimization further, FastServe\cite{fastserve} enhances optimization through the careful consideration of service latency. It employs a preemptive scheduling mechanism alongside an innovative skip-join Multi-Level Feedback Queue scheduler. This combined approach aims to curtail job completion time and diminish both requests waiting and processing durations. Moreover, the skip-join mechanism anticipates the time required for the initial iteration, thus swiftly delegating the job to the most suitable queue. This strategic allocation aids in averting unnecessary queue transitions and subsequent delays. However, despite these sophisticated approaches, Current methods\cite{SHEPHERD-c1,SHEPHERD-c2}  that make decisions for each request individually often result in GPU over-provisioning during short-term high loads, leading to low resource utilization. To address this issue, Shepherd\cite{SHEPHERD} improves predictability by grouping unpredictable individual request streams into medium-sized batches. It employs a two-stage scheduling algorithm. In the first stage, it utilizes long-term load data to divide the GPU cluster into service groups and allocate GPUs accordingly. In the second stage, it introduces preemptive scheduling, prioritizing larger batches that align with SLOs to optimize throughput even with reactive scheduling.

In scenarios with strong latency requirements, some works have addressed the issue. Clockwork\cite{clockwork} ensures predictable DNN inference times, combating tail latency from diverse tasks, hardware, and inputs to satisfy SLOs and minimize delays. By limiting options at each computational layer for uniform execution times and deploying a central controller that assigns tasks with known durations, Clockwork maintains strict timing assurances. REEF\cite{reef} is a system designed for efficient and timely DNN inference on GPUs. It schedules tasks in a way that prioritizes real-time tasks, quickly interrupts other tasks if needed, and allocates computing units first to real-time kernels, then distributes the remaining units to best-effort kernels. In the field of latency-sensitive services, OSML\cite{osml} is a machine learning-based scheduler that integrates architectural metrics like IPC and cache misses into a predictive model for Quality of Service (QoS) changes. It deploys three models: Model A calculates optimal resource allocations and detects ``resource cliffs"; Model B redistributes resources, prioritizing services that are sensitive to QoS degradation; and Model C dynamically adjusts resource allocation in real-time based on ongoing QoS assessments to maintain service performance.

\subsection{GPU Memory Optimization In Inference}
During the process of inference, the weight parameters of a model significantly consume GPU memory. Various studies have concentrated on optimizing these model parameters. For instance, FlexGen\cite{flexgen} is a throughput-oriented generative inference system that optimizes offloading strategies. A standout characteristic of FlexGen is its zig-zag block scheduling strategy. The zig-zag block scheduling strategy explores the computation graph by advancing column-by-column and reusing weights within each column to minimize loading times. When the memory limits for activations are reached, the process transitions to the next column, optimizing GPU memory utilization and efficiently processing the model through a zig-zag pattern. Additionally, it dynamically loads and unloads activation values and KV Cache as required. In another study, Jeong et al.\cite{dha} utilized Direct-Host-Access (DHA) for direct GPU memory access, reducing latency for layers like the embedding layer. They also applied Parallel Model Transmission, dividing the model per GPU for parallel loading via PCIe. The sections are then quickly transferred to the primary GPU using NVLink, optimizing layer execution.

GPU memory constraints hinder foundation model inference, where the storage of the KV Cache is a significant memory overhead. Serving architectures use KV Cache to reduce re-computation, but as the token count grows, so does the cache, risking GPU memory overflow. To avoid this, frameworks limit iteration length and pre-allocate memory for KV Cache, which can lead to memory fragmentation and reduced inference performance. Several techniques have been proposed to optimize this aspect. The PageAttention mechanism proposed by vLLM\cite{vllm} addresses the issues of GPU memory over-allocation and fragmentation. It accomplishes this by emulating OS page table mapping and segmenting GPU memory into blocks. A block mapping table is then used to ensure logically sequential but physically discrete storage. This dynamic approach effectively meets the demand of the KV Cache, reducing memory fragmentation and improving inference throughput. Drawing inspiration from the virtual nature of operating systems, The gpulet\cite{gpulet} concept introduces an abstraction for partitioning GPUs, creating virtual GPUs that possess a fraction of the physical GPU resources. The proposed multidimensional search-based scheduling framework optimizes GPU tasks by considering data batch sizes along with the temporal and spatial sharing of resources.

\subsection{Multi-Model Inference}
Multi-model inference involves utilizing multiple models for serving. These models can be of the same type or different types, often with varying architectures. An important research question in this context is how to effectively combine these diverse models and optimize resource allocation to achieve optimal performance. In the context of multi-task models created through fine-tuning, PetS\cite{pets} introduces an innovative framework for multi-task Parameter Efficient Transformers (PET) that processes various tasks in a unified manner. Traditional fine-tuning of the entire model for each task incurs substantial memory overhead. PetS circumvents this by employing parameter-efficient fine-tuning methods such as Adapters\cite{adapter2,adapter1}, splitting the model into a shared core and task-specific small operators. This architecture allows for shared base model usage across tasks, reducing memory demands and streamlining model deployment. In the context of hierarchical models ranging from small to large, one approach is presented by Tabi\cite{tabi}. Tabi leverages the observation that smaller models often exhibit predictive capabilities similar to larger models by implementing a multi-level inference engine. It employs well-calibrated confidence scores using temperature scaling to determine whether a query can be promptly resolved using the smaller model or if it should be escalated to the larger model. For escalated queries, Tabi reduces system overhead by employing attention-based word pruning and a weighted ensemble approach. Another technique introduced by Google Research is Speculative Decoding\cite{speculativedecoding}, which aims to accelerate the inference process for language models. This method involves a smaller model generating tokens sequentially while a larger model simultaneously validates the correctness of each token in parallel. The larger model verifies the sequence of tokens produced by the smaller model, enabling the generation of multiple tokens within a single iteration of the larger model. LLMCad\cite{llmcad} differs from Google's Speculative Decoding by employing a tree-based token generation approach that facilitates the concurrent evaluation of multiple tokens. To accomplish this, LLMCad utilizes a smaller language model to construct a comprehensive vocabulary tree comprising various word paths. The larger LLM then efficiently and concurrently evaluates these paths.

\section{CHALLENGE AND FUTURE DIRECTIONS}
\ding{202} \noindent \textbf{Privacy protection.} 
Regarding privacy protection, the key challenge for foundation models lies in the potential unauthorized collection, usage, and inadvertent disclosure of personal information. Future efforts should focus on incorporating privacy protection mechanisms into the design and application of models to ensure robust safeguards for user data, preventing unauthorized use and disclosure threats.

\noindent \ding{203} \noindent \textbf{Security.} Foundation models exhibit a relatively weak ability to defend against malicious attacks, making them susceptible to activities such as command injection and prompt injection. Particularly in critical domains such as politics, military, finance, and healthcare, any form of malicious attack could severely affect the stability of national society and the safety of people's lives and property. Therefore, future efforts must focus on enhancing security measures for foundation models to ensure their reliable protection in critical domains.

\noindent \ding{204} \noindent \textbf{Energy
sustainability.} Foundation systems face a significant challenge in terms of energy sustainability during both training and serving. This entails a high demand for substantial computational resources, which may result in adverse environmental impacts. The key to future efforts lies in enhancing the energy efficiency of models and adopting more energy-efficient hardware innovations. Through innovative green computing and sustainable development, these efforts aim to make foundation model systems more environmentally friendly and efficient, reducing energy dependence and mitigating environmental impact.

\section{CONCLUSION}
This survey delves into the training and serving methods of foundation model systems from the perspectives of network, computing, and storage.
In the training section, it discusses various parallel computing strategies. Each strategy has unique advantages and application scenarios. Additionally, it explores GPU memory optimization and communication optimization techniques.
The serving section discusses key technologies such as batch processing, sparse acceleration, resource scheduling, GPU memory optimization, and multi-model
inference. These strategies are essential for ensuring the efficiency and practicality of the foundation model system in real-world scenarios.
In summary, the training and serving of foundation model systems is an evolving field. With the emergence of new technologies, it anticipates solving more challenges and further advancing the field of artificial general intelligence.
\section*{REFERENCES}

\def\refname{\vadjust{\vspace*{-1em}}} %Please don't do this in a real paper.

\end{document}